\pdfoutput=1

\documentclass[11pt]{article}

\usepackage[final]{acl2023}

\usepackage{times}
\usepackage{latexsym}
\usepackage{amsmath, amssymb}
\usepackage{hyperref}
\usepackage{graphicx}
\usepackage{cite}
\usepackage{geometry}
\usepackage[most]{tcolorbox}

\usepackage[T1]{fontenc}

\usepackage[utf8]{inputenc}

\usepackage{microtype}

\usepackage{inconsolata}
\usepackage{graphicx,adjustbox,mathtools,varwidth}
\usepackage{amsmath}      
\usepackage{booktabs}     
\usepackage{graphicx}     
\usepackage{array}        
\usepackage{multirow}     
\usepackage{caption}      
\usepackage{algorithm}
\usepackage{algorithmic}

\title{AGSC: Adaptive Granularity and Semantic Clustering for Uncertainty Quantification in Long-text Generation}

\author{
Guanran Luo\textsuperscript{\textdagger},
Wentao Qiu\textsuperscript{\textdagger},
Wanru Zhao,
Wenhan Lv,
Zhongquan Jian,\\
\textbf{Meihong Wang},
\textbf{Qingqiang Wu}\\
School of Informatics, Xiamen University\\
\texttt{luoguanran@stu.xmu.edu.cn, wuqq@xmu.edu.cn}
}

\begin{document}
\maketitle

\begingroup
\renewcommand\thefootnote{}
\footnotetext{\textsuperscript{\textdagger} These authors contributed equally to this work.}
\endgroup

\begin{abstract}
Large Language Models (LLMs) have demonstrated impressive capabilities in long-form generation, yet their application is hindered by the hallucination problem. While Uncertainty Quantification (UQ) is essential for assessing reliability, the complex structure makes reliable aggregation across heterogeneous themes difficult, in addition, existing methods often overlook the nuance of neutral information and suffer from the high computational cost of fine-grained decomposition. To address these challenges, we propose \textbf{AGSC} (\textbf{A}daptive \textbf{G}ranularity and \textbf{G}MM-based \textbf{S}emantic \textbf{C}lustering), a UQ framework tailored for long-form generation. AGSC first uses NLI neutral probabilities as triggers to distinguish \textit{irrelevance} from \textit{uncertainty}, reducing unnecessary computation. It then applies Gaussian Mixture Model (GMM) soft clustering to model latent semantic themes and assign topic-aware weights for downstream aggregation. Experiments on BIO and LongFact show that AGSC achieves state-of-the-art correlation with factuality while reducing inference time by about 60\% compared to full atomic decomposition.
\end{abstract}

\section{Introduction}

With the rapid development and widespread application of Large Language Models (LLMs), their potential in generating long-text content is increasingly prominent. Traditional Natural Language Processing (NLP) tasks have evolved into use cases primarily involving the generation of long-text responses \citep{zhao2023llmsurvey,chang2023llmeval}. However, a core challenge for LLMs lies in their inherent hallucination problem, where the model may output inaccurate or unfaithful information with high confidence \citep{manakul2023selfcheckgpt,zhang2023hallucinationsurvey,wang2023factuality}. Therefore, accurate and robust Uncertainty Quantification (UQ) of LLM outputs is essential for enhancing their trustworthiness and application in critical domains.

Previous research on modeling uncertainty has mainly focused on short responses \citep{kuhn2023semantic,duan2023sar,lin2024generating,xiao2022uqplm,xiong2023llmuncertainty,huang2023lookbeforeyouleap}. However, with the widespread application of Large Language Models (LLMs), generative tasks involving long text have become more common use cases, such as summarization, open-ended question answering, and multi-agent collaboration. Traditional confidence-based and semantic entropy methods \citep{hendrycks2017baseline,geifman2017selective,lin2022express,malinin2021autoregressive,mielke2022linguisticcalibration} can only make a single overall correctness judgment for the entire response, failing to capture local errors in different facts within long text.

\begin{figure}[t]
    \centering
    \includegraphics[width=\linewidth]{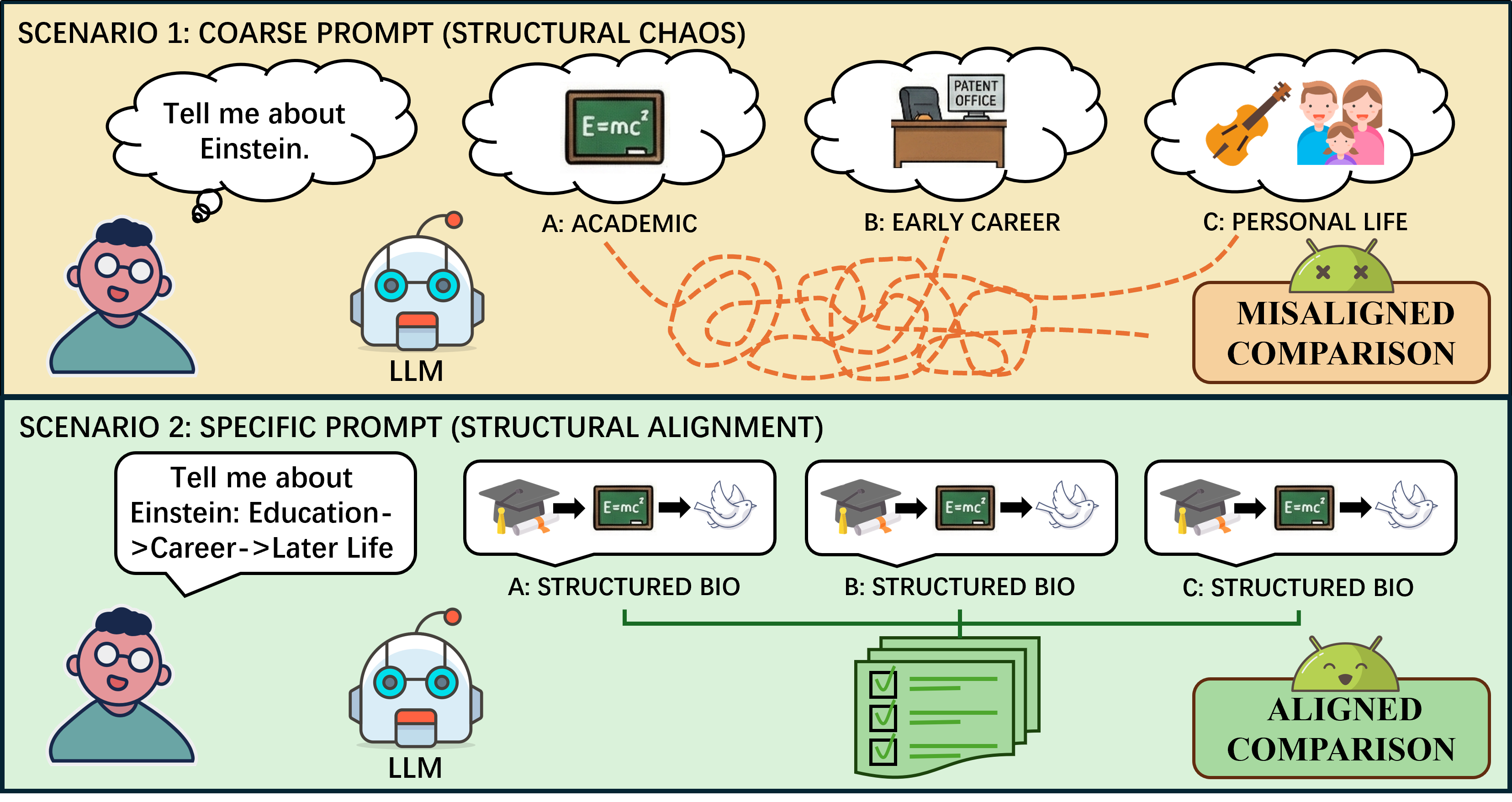}
    \caption{Illustration of prompt-induced structural uncertainty.}
    \label{fig:prompt_variation}
\end{figure}

LUQ \citep{zhang2024luq} addresses this by breaking down each response into sentences or atomic facts as the hypothesis and checking if they can be supported by other samples serving as the premise, thus enabling a more granular evaluation. However, this method has the following limitations:

\begin{enumerate}
  \item \textbf{Efficiency-Granularity Trade-off:} 
  Fine-grained decomposition drastically increases computational overhead and LLM calls, making it difficult to balance accuracy with efficiency.

 \item \textbf{Topic Heterogeneity in Aggregation:}
 Long-form responses often mix multiple semantic themes, where minor/off-topic parts can disproportionately affect overall uncertainty if all units are pooled uniformly.
 A theme-aware aggregation mechanism is needed to down-weight such minor/noisy portions.

  \item \textbf{Underutilization of Neutrality:} 
  LUQ simply discards the \emph{neutral} label. However, neutrality often signals epistemic uncertainty that requires further granularity analysis rather than direct exclusion.
\end{enumerate}

Figure~\ref{fig:prompt_variation} provides an intuitive explanation.
With a coarse prompt such as ``Tell me about Einstein,'' different samples may organize the content around different latent aspects (e.g., academic vs.\ early career vs.\ personal life), leading to \emph{structural chaos}.
As a result, different samples may emphasize different themes, and naive pooling can be overly influenced by minor or off-topic parts.
Moreover, coarse prompts often introduce many contextually irrelevant sentences; when compared against a given hypothesis sentence, these irrelevant contexts tend to yield a high probability of the \emph{neutral} class.

To address these limitations, we propose \textbf{AGSC} (\textbf{A}daptive \textbf{G}ranularity and \textbf{S}emantic \textbf{C}lustering). AGSC leverages the NLI neutral category to dynamically trigger decomposition for accurate yet efficient evaluation, and employs Gaussian Mixture Model (GMM) soft clustering to capture latent semantic themes and perform topic-weighted aggregation, reducing the influence of minor/noisy parts.
Finally, uncertainty is computed from NLI-based unit uncertainties and aggregated with cluster-aware weights.
Experiments on FActScore \citep{min2023factscore} and LongFact \citep{wei2024longfact} show that AGSC achieves state-of-the-art (SOTA) correlations with factuality. 

Our main contributions are:
\begin{itemize}
    \item We propose an \textbf{Adaptive Granularity Strategy} that uses NLI neutrality to trigger fine-grained decomposition or filter noise, effectively balancing UQ accuracy with computational efficiency.
    
    \item We introduce \textbf{GMM-based Semantic Clustering} to model latent themes and assign topic-aware weights, which down-weights minor/noisy parts during aggregation and yields more stable long-form uncertainty estimates.

    \item AGSC achieves \textbf{SOTA performance} on two long-text benchmarks, consistently outperforming baselines in correlating uncertainty with factuality.
\end{itemize}

\section{Related Work}
\label{sec:related_work}

\begin{figure*}[ht] \centering \includegraphics[width=0.95\textwidth]{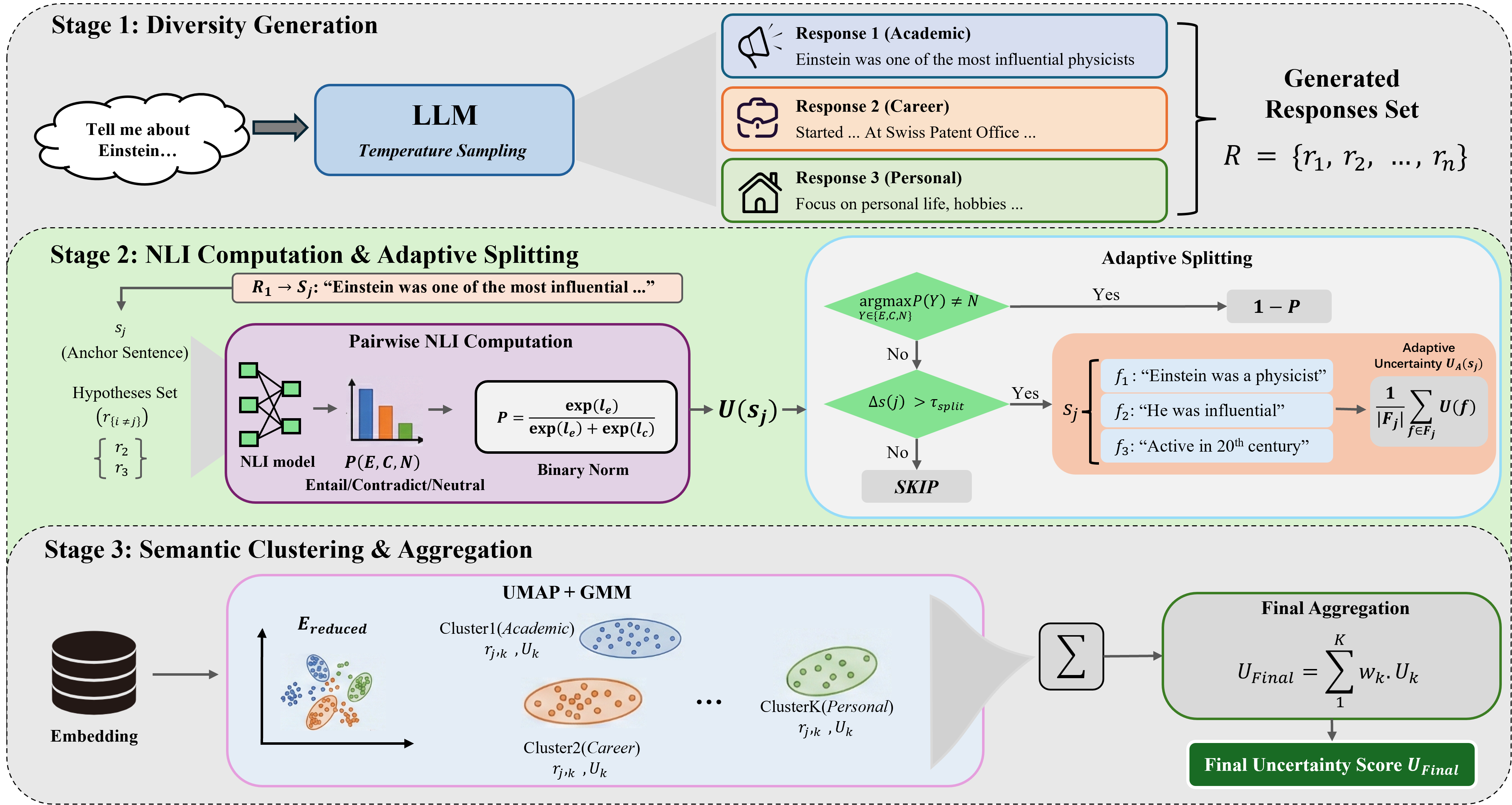} \caption{Overview of the \textbf{AGSC} framework. The framework consists of three stages: (1) Diversity Generation, where multiple responses are sampled; (2) NLI Computation \& Adaptive Splitting, where sentences are analyzed via NLI and adaptively decomposed into atomic facts or skipped; and (3) Semantic Clustering \& Aggregation, where units are softly clustered via UMAP and GMM to derive the final uncertainty score.}
\label{fig:framework} \end{figure*}

\paragraph{Epistemic Uncertainty in Machine Learning.}
Uncertainty quantification has been a foundational topic in machine learning long before the era of LLMs \citep{gawlikowski2023uncertainty,zhang2024adacomp,zhang2025less,wang2026perm,zhang2026stable,wang2025think}. Standard taxonomies categorize uncertainty into two primary types: \textit{aleatoric uncertainty} and \textit{epistemic uncertainty} \citep{hora1996aleatory,hullermeier2021aleatoric,der2009aleatoric}.
In the context of LLM hallucination, non-factual generation is often attributed to the model's lack of relevant knowledge \citep{huang2023lookbeforeyouleap}. Therefore, consistent with recent work \citep{kuhn2023semantic,lin2024generating}, our research focuses on \textbf{epistemic uncertainty}. Recent studies have also explored learning conditional dependencies to disentangle unconditional truthfulness from generation artifacts \citep{vazhentsev2025unconditional}, further refining the theoretical basis for generative UQ.

\paragraph{Uncertainty Quantification in LLMs.}
Existing approaches are generally divided into white-box and black-box methods \citep{geng2023confidencecalibration}.
\textit{White-box methods} rely on internal logits or attention. Measures such as Semantic Entropy (SE) \citep{kuhn2023semantic} use token probabilities to estimate confidence. Complementary formulations for in-context learning scenarios have also been proposed to better leverage model internals \citep{ling2024uncertainty}. However, these methods often fail to capture semantic equivalence in long sequences \citep{duan2023sar} and are inapplicable to closed-source models.
\textit{Black-box methods} have thus gained prominence. \textit{Verbalized uncertainty} prompts models to express confidence \citep{lin2022express,tian2023just}, though recent work on "self-known" vs. "self-unknown" knowledge suggests these self-assessments can be intricate and prone to miscalibration \citep{tu2025investigating}.
\textit{Consistency-based methods}, which AGSC follows, measure semantic divergence among sampled responses \citep{manakul2023selfcheckgpt,wang2023factuality}. Beyond simple sampling, perturbation-based methods like SPUQ \citep{gao2024spuq} introduce input noise to assess stability, offering a robust alternative to evaluating epistemic uncertainty in black-box settings.

\paragraph{Granularity and Semantic Clustering.}
A critical open problem in long-form UQ is the choice of \textit{granularity} and how to aggregate evidence under
\textit{topic heterogeneity}. Early works operated at the sentence level \citep{manakul2023selfcheckgpt}, which may miss
mixed veracity within long sentences. Recent methods move towards finer granularity by decomposing responses into
\textit{atomic facts} \citep{min2023factscore,zhang2024luq}, but this often incurs substantial computational overhead. 
Beyond granularity, long responses may mix multiple semantic themes; naive uniform pooling can be overly influenced by
minor or off-topic portions. Several lines of work leverage semantic similarity or structured representations to improve
fine-grained UQ, such as Kernel Language Entropy \citep{nikitin2024kernel} and graph-based formulations \citep{lin2024generating}.
KG-based decomposition \citep{yuan2025kguq} can offer precise structure but depends on external extractors.
In contrast, AGSC focuses on (i) adaptively selecting granularity using NLI neutrality signals to reduce unnecessary
decomposition, and (ii) applying GMM-based \emph{soft semantic clustering} to obtain theme-aware weights, enabling more
stable aggregation in long-form uncertainty estimation.

\section{Method}

In this section, we propose \textbf{AGSC} (Adaptive Granularity and Semantic Clustering). As illustrated in Figure~\ref{fig:framework}, our method consists of three primary stages: (1) \textbf{Adaptive Granularity Triggering}; (2) \textbf{GMM-based Semantic Clustering}; and (3) \textbf{Uncertainty Aggregation}.

\subsection{Notation}
First, we define a terminology hierarchy:
\begin{itemize}
    \item \textbf{Text Unit (Unit):} The general term for any segment being evaluated.
    \item \textbf{Sentence:} The coarse-grained unit from the original generation.
    \item \textbf{Atomic Fact:} The fine-grained unit resulting from decomposition.
    \item \textbf{Premise Unit ($u$):} Refers to the specific Text Unit (Sentence or Atomic Fact) currently serving as the premise for NLI verification.
\end{itemize}

We sample $n$ responses $\mathcal{R}=\{r_1,\dots,r_n\}$ via temperature sampling, and treat $r_1$ as the \emph{anchor response} to be evaluated.
The remaining responses $\mathcal{R}_{\mathrm{ref}}=\{r_2,\dots,r_n\}$ serve as \emph{reference evidence}. After adaptive granularity (Sec.~\ref{sec:adaptive-granularity}),
each response yields a set of valid semantic units $\mathcal{H}_i^*$. Each response $r_i$ is segmented into sentences $\mathcal{S}_i=\{s_{i,1},\dots,s_{i,|\mathcal{S}_i|}\}$.
We denote the union of all sentences as $\mathcal{S}=\bigcup_{i=1}^{n}\mathcal{S}_i$.
After clustering (Sec.~\ref{sec:gmm-alignment}), each sentence $s_{i,j}$ is associated with a soft membership
vector $\boldsymbol{\gamma}_{i,j}\in\mathbb{R}^{K}$, where $\gamma_{i,j,k}$ denotes its posterior responsibility
to cluster $k$ and $\sum_{k=1}^{K}\gamma_{i,j,k}=1$.

For each NLI pair, let $l_e$, $l_c$, and $l_n$ denote the raw logits for the entailment, contradiction, and neutral classes, respectively.
We use $P_{\text{3cls}}(\cdot)$ to denote the original three-way NLI distribution over $\{\texttt{entail}, \texttt{contradict}, \texttt{neutral}\}$, and $P_{\text{binary}}(\cdot)$ to denote the binary-normalized distribution over $\{\texttt{entail}, \texttt{contradict}\}$ after removing the neutral class.

\subsection{NLI-based Uncertainty Computation}
\label{sec:nli-uncertainty}

\paragraph{Pairwise NLI score.} 
For a premise unit $u$ (a sentence or atomic fact from the anchor) and a hypothesis text $v$ (a reference response), we employ a sliding window strategy to address the context length limit of NLI models. 
Specifically, the hypothesis $v$ is segmented into a set of overlapping chunks $\mathcal{C}_v = \{c^{(1)}, \dots, c^{(m)}\}$. 
We compute the NLI logits for $u$ paired with each chunk individually. 
To ensure robust support detection, we aggregate these scores by selecting the chunk that yields the maximum entailment probability, assuming that valid support in any segment suffices to entail the premise.

Based on the logits $(l_e, l_c, l_n)$ of this best-matching chunk, we ignore the neutral class in a binary normalization. The operation $\max$ is taken over all sliding window chunks $k \in \{1, \dots, m\}$ of the reference text $v$:
\begin{equation}
    \resizebox{\linewidth}{!}{$
    P(\text{entail}|u,v) = \max_{k \in \{1, \dots, m\}} \left( \frac{\exp(l_e^{(k)})}{\exp(l_e^{(k)}) + \exp(l_c^{(k)})} \right)
    $}
    \label{eq:binary_norm}
\end{equation}
In the context of long-text hallucination detection, ``neutral'' sentences typically represent irrelevant information or chit-chat that does not impact the overall veracity of the response. Therefore, removing the neutral score serves to amplify the semantic tendency (Entailment vs. Contradiction) of the content without reversing its polarity.

\paragraph{Anchor-to-reference entailment support and uncertainty.}
Inspired by the entailment-support-based uncertainty principle in long-form UQ \citep{zhang2024luq},
we define the entailment support for an anchor unit $h\in\mathcal{H}_1^*$ as the average entailment score
against all whole reference responses:
\begin{equation}
    S(h) = \frac{1}{n-1}\sum_{t=2}^{n} P(\text{entail}|h,r_{t}), \quad h \in \mathcal{H}_{1}^{*}
\end{equation}
Then the uncertainty of an anchor unit is defined as:
\begin{equation}
U(h) = 1 - S(h),
\qquad h\in\mathcal{H}_1^*.
\label{eq:unit-uncertainty}
\end{equation}

\subsection{Adaptive Granularity}
\label{sec:adaptive-granularity}

Neutral NLI predictions may reflect either (i) an \textit{uncertainty tendency} (ambiguous content that should be decomposed)
or (ii) an \textit{irrelevance tendency} (off-topic noise that should be skipped). We distinguish them using the
entailment--contradiction gap computed from anchor-to-reference NLI signals.
For a sentence $s_j$ in the anchor response $r_1$, we compute its average NLI distribution across whole reference responses:
\begin{equation}
\bar{P}\!\left(Y \mid s_j\right)
=
\frac{1}{n-1}\sum_{t=2}^{n} P\!\left(Y \mid s_j, r_t\right),
\label{eq:avg-nli}
\end{equation}
and define the dominant label:
\begin{equation}
\hat{y}_j
=
\operatorname*{argmax}_{Y \in \{\text{entail}, \text{contra}, \text{neutral}\}}
\bar{P}\!\left(Y \mid s_j\right).
\end{equation}

\paragraph{Decompose vs.\ skip.}
If $\hat{y}_j \neq \text{neutral}$, we keep $s_j$ and later compute $U(s_j)$ using Eq.~\ref{eq:unit-uncertainty}.
If $\hat{y}_j=\text{neutral}$, we compute the entailment--contradiction gap:
\begin{equation}
\Delta(s_j)=\big|\bar{P}(\text{entail}\mid s_j)-\bar{P}(\text{contra}\mid s_j)\big|.
\end{equation}
If $\Delta(s_j)>\tau$, we decompose $s_j$ into atomic facts $F_j=\mathrm{Split}(s_j)$.
Otherwise, $s_j$ is treated as irrelevance noise and marked as \textsc{Skip}.

\paragraph{Adaptive uncertainty.}
The adaptive uncertainty is:
\begin{equation}
\resizebox{0.95\columnwidth}{!}{$
  U_{\text{A}}(s_j)
  =
  \begin{cases}
    U(s_j), & \hat{y}_j \neq \text{neutral}, \\[4pt]
    \textsc{Skip}, & \hat{y}_j = \text{neutral} \land \Delta(s_j) \le \tau, \\[6pt]
    \dfrac{1}{\lvert F_j \rvert}
    \displaystyle\sum_{f\in F_j}
    U(f), & \hat{y}_j = \text{neutral} \land \Delta(s_j) > \tau,
  \end{cases}
$}
\end{equation}
where $U(\cdot)$ is computed by Eq.~\ref{eq:unit-uncertainty}. Units marked as \textsc{Skip} are excluded from
subsequent aggregation.

\subsection{GMM-Based Semantic Clustering}
\label{sec:gmm-alignment}

Long-form generations may contain multiple semantic themes. To reduce the influence of minor/noisy parts during
aggregation, we perform soft clustering over sentence units from all responses and use the resulting
memberships as theme-aware weights for downstream aggregation.

\paragraph{Embedding and dimensionality reduction.}
Let $\mathcal{H}^*=\bigcup_{i=1}^{n}\mathcal{H}_i^*$ denote the union of all valid semantic units after adaptive
granularity. We encode each unit $h\in\mathcal{H}^*$ into a sentence embedding $e_h \in \mathbb{R}^{D}$ using a
pretrained embedding model, and denote $E=\{e_h\}_{h\in\mathcal{H}^*}$.

To improve robustness and efficiency for clustering, we optionally apply UMAP \citep{mcinnes2018umap} to project
embeddings from $\mathbb{R}^{D}$ to $\mathbb{R}^{D'}$ ($D'\ll D$), yielding reduced embeddings
$E_{\text{reduced}}=\{e_h^{\text{reduced}}\in\mathbb{R}^{D'}\}$.

\paragraph{Soft clustering via GMM.}
On the reduced embeddings $\{e_h^{\text{reduced}}\}$, we fit a $K$-component GMM \citep{bishop2006prml}.
The generative model assumes a latent cluster variable $z_h \in \{1,\dots,K\}$:
\begin{equation}
\resizebox{\linewidth}{!}{$
p(z_h=k)=\pi_k,\qquad
p\!\left(e_h^{\text{reduced}}\mid z_h=k\right)=\mathcal{N}\!\left(e_h^{\text{reduced}}\mid \mu_k,\Sigma_k\right).
$}
\label{eq:gmm_generative}
\end{equation}
Parameters are learned by maximizing the log-likelihood:
\begin{equation}
\resizebox{\linewidth}{!}{$
\log p(E_{\text{reduced}})
=
\sum_{h\in\mathcal{H}^*}
\log\left(\sum_{k=1}^K \pi_k\,\mathcal{N}\!\left(e_h^{\text{reduced}}\mid \mu_k,\Sigma_k\right)\right).
$}
\label{eq:gmm_ll}
\end{equation}

For each unit $h$, the posterior probability of belonging to cluster $k$ serves as its soft semantic weight:
\begin{equation}
\resizebox{\linewidth}{!}{$
\gamma_{h,k}
=
p(z_h=k\mid e_h^{\text{reduced}})
=
\frac{\pi_k\,\mathcal{N}\!\left(e_h^{\text{reduced}}\mid \mu_k,\Sigma_k\right)}
{\sum_{\ell=1}^K \pi_\ell\,\mathcal{N}\!\left(e_h^{\text{reduced}}\mid \mu_\ell,\Sigma_\ell\right)}.
$}
\label{eq:gmm_posterior}
\end{equation}

Intuitively, unlike hard clustering methods which classify points as either belonging to a single cluster or noise---thereby discarding necessary fine-grained probability weights---GMM provides the posterior probability $P(z=k|x)$ (soft assignment). This probabilistic membership is essential for handling semantically ambiguous sentences in long-form generation, better matching the ambiguous boundaries of natural language semantics and providing smoother, more stable aggregation downstream.

\subsection{Uncertainty Aggregation}
\label{sec:aggregation}

\paragraph{Cluster mass and cluster uncertainty.}
We define the mass of cluster $k$ as the accumulated membership probability over anchor units:
\begin{equation}
M_k = \sum_{h \in \mathcal{H}_1^*} \gamma_{h,k}.
\end{equation}
The cluster-specific uncertainty is the mass-normalized average of anchor-unit uncertainties:
\begin{equation}
U_k
=
\frac{1}{M_k}
\sum_{h \in \mathcal{H}_1^*} \gamma_{h,k}\,U(h).
\label{eq:cluster-uncertainty}
\end{equation}

\paragraph{Mass-weighted global uncertainty.}
We set $w_k = M_k / \sum_{j=1}^{K} M_j$ so dominant themes in the anchor response contribute more.
Finally, the uncertainty score for the anchor response $r_1$ is:
\begin{equation}
U_{\mathrm{Final}} = \sum_{k=1}^{K} w_k \cdot U_k.
\end{equation}
\section{Experimental Setup}
\label{sec:experimental_setup}

\subsection{Datasets, Metrics, and Baselines}

\noindent\textbf{Datasets.} We evaluate our method on two widely recognized benchmarks for long-form generation: \textbf{BIO} \citep{min2023factscore}, which focuses on biography generation, and \textbf{LongFact} \citep{wei2024longfact}, a comprehensive dataset covering diverse topics for long-context factuality evaluation.

For BIO, we use a test set of 500 entities, and generate 5 biographies per entity.
For LongFact, we sample 10 prompts for each of 38 topics, yielding 380 prompts in total, and generate 5 responses per prompt.

\vspace{0.5em}
\noindent\textbf{Metrics.} We employ \textbf{FActScore} \citep{min2023factscore} as the ground-truth measure for the factuality of generated responses. Specifically, we apply FActScore to the \textbf{first generated response} for each question. To assess the performance of Uncertainty Quantification (UQ) methods, we measure how well the estimated uncertainty scores align with the actual factuality. Following standard practice for correlation analysis \citep{schober2018correlation}, we report the \textbf{Pearson Correlation Coefficient (PCC)} and \textbf{Spearman Correlation Coefficient (SCC)} between the calculated uncertainty scores and the FActScore.

\vspace{0.5em}
\noindent\textbf{Models.} We use \textbf{GPT-4.1-mini} \citep{openai2025gpt41}, \textbf{Qwen2.5-32B} \citep{bai2023qwen}, and \textbf{Llama3-70B} \citep{meta2024llama3} as the generator models.
For the uncertainty estimation components, we utilize \textbf{DeBERTa-v3-large-mnli} \citep{he2021deberta} as the Natural Language Inference (NLI) model to predict entailment relationships. For semantic clustering, we employ \textbf{gte-large-en-v1.5} \citep{li2023gte} to generate sentence embeddings.

\vspace{0.5em}
\noindent\textbf{Baselines.} We compare our approach against a comprehensive set of baselines: token-level \textbf{SE} (Semantic Entropy) \citep{kuhn2023semantic}; similarity-based \textbf{LexSim} (Lexical Similarity); graph-based measures including \textbf{Ecc} (Eccentricity) and \textbf{Deg} (Degree Matrix) \citep{lin2024generating}. We also include advanced semantic baselines: \textbf{KLE} (Kernel Language Entropy) \citep{nikitin2024kernel}, which estimates fine-grained uncertainty via semantic similarities; \textbf{SPUQ} \citep{gao2024spuq}, a perturbation-based method for robust uncertainty estimation; and the NLI-based \textbf{SCN} (SelfCheckNLI) \citep{manakul2023selfcheckgpt}. We also include SAR (Shifting Attention to Relevance) \citep{duan2024sar}, a relevance-aware uncertainty estimation method. Finally, we compare with the \textbf{LUQ} framework and its variants \textbf{LUQ-Pair} and \textbf{LUQ-Atomic} \citep{zhang2024luq}.

\subsection{Implementation Details}
All experiments were conducted on an \textbf{NVIDIA GeForce RTX 4090} GPU. We generated $n=5$ responses for each question using temperature sampling ($t=0.7$).

For semantic clustering, we implement an adaptive soft clustering algorithm to obtain \emph{theme-aware membership weights}
used in downstream aggregation. We first apply \textbf{UMAP} to reduce embedding dimensions to \textbf{32}, configured with
\textbf{15} nearest neighbors, a minimum distance of \textbf{0.1}, \textbf{cosine} metric, and \textbf{spectral}
initialization. Then, we employ a GMM with \textbf{full} covariance type, \textbf{k-means++} initialization, and a
regularization constant of \textbf{$10^{-5}$}. The optimal number of clusters $K$ is determined dynamically based on the
\textbf{Bayesian Information Criterion (BIC)} with an improvement threshold of \textbf{0.01}. The pseudocode for this
algorithm is provided in \textbf{Appendix~\ref{app:clustering_algorithm}}.

For adaptive granularity triggering, we set the entailment--contradiction gap threshold to $\tau=0.1$.

To decompose long-form responses into atomic facts, we utilize the \texttt{AtomicFactGenerator} from the \textbf{FActScore} library \citep{min2023factscore}. Following the standard implementation, this module employs a few-shot prompting strategy to split sentences into independent facts.
\section{Results}

\subsection{Main Results} \label{sec:main_results}

\begin{table*}[t]
\centering
\resizebox{\linewidth}{!}{%
\begin{tabular}{llcccccccccccccc}
\toprule
\textbf{Dataset} & \textbf{Model} & \textbf{Metric $\downarrow$} & SE & LexSim & Ecc & Deg & KLE & SPUQ & SCN & SAR & LUQ & LUQ-Pair & LUQ-Atomic & \textbf{AGSC} \\
\midrule
\multirow{6}{*}{\textbf{BIO}} & \multirow{2}{*}{GPT-4.1-mini} & PCC & - & -0.438 & -0.219 & -0.329 & -0.501 & -0.515 & -0.493 & -0.515 & -0.548 & -0.381 & \underline{-0.572} & \textbf{-0.708} \\
 & & SCC & - & -0.423 & -0.211 & -0.317 & -0.495 & -0.508 & -0.486 & 	-0.490 & \underline{-0.529} & -0.241 & -0.486 & \textbf{-0.665} \\
\cmidrule{2-15}
 & \multirow{2}{*}{Qwen2.5-32B} & PCC & -0.246 & -0.262 & -0.131 & -0.197 & -0.305 & -0.315 & -0.295 & -0.405 & -0.328 & -0.414 & \underline{-0.517} & \textbf{-0.645} \\
 & & SCC & -0.318 & -0.339 & -0.169 & -0.254 & -0.390 & -0.405 & -0.382 & -0.450 & -0.424 & -0.395 & \underline{-0.541} & \textbf{-0.646} \\
\cmidrule{2-15}
 & \multirow{2}{*}{Llama3-70B} & PCC & -0.242 & -0.258 & -0.129 & -0.193 & -0.295 & -0.305 & -0.290 & -0.285 & \underline{-0.322} & -0.255 & -0.239 & \textbf{-0.339} \\
 & & SCC & -0.251 & -0.268 & -0.134 & -0.201 & -0.308 & -0.315 & -0.301 & -0.290 & \underline{-0.334} & -0.260 & -0.250 & \textbf{-0.335} \\
\midrule
\multirow{6}{*}{\textbf{LongFact}} & \multirow{2}{*}{GPT-4.1-mini} & PCC & - & -0.163 & -0.082 & -0.122 & -0.190 & -0.195 & -0.184 & -0.255 & -0.204 & -0.324 & \underline{-0.327} & \textbf{-0.370} \\
 & & SCC & - & -0.167 & -0.084 & -0.125 & -0.195 & -0.201 & -0.188 & -0.280 & -0.209 & -0.298 & \underline{-0.445} & \textbf{-0.461} \\
\cmidrule{2-15}
 & \multirow{2}{*}{Qwen2.5-32B} & PCC & -0.131 & -0.140 & -0.070 & -0.105 & -0.162 & -0.168 & -0.158 & -0.210 & -0.175 & \underline{-0.288} & -0.255 & \textbf{-0.291} \\
 & & SCC & -0.089 & -0.095 & -0.048 & -0.071 & -0.112 & -0.115 & -0.107 & -0.145 & -0.119 & \underline{-0.203} & -0.175 & \textbf{-0.256} \\
\cmidrule{2-15}
 & \multirow{2}{*}{Llama3-70B} & PCC & -0.032 & -0.034 & -0.017 & -0.025 & -0.040 & -0.041 & -0.038 & -0.095 & -0.042 & \textbf{-0.178} & -0.128 & \underline{-0.175} \\
 & & SCC & -0.067 & -0.071 & -0.036 & -0.053 & -0.082 & -0.085 & -0.080 & -0.110 & -0.089 & -0.069 & \underline{-0.169} & \textbf{-0.229} \\
\bottomrule
\end{tabular}%
}
\caption{Pearson (PCC) and Spearman (SCC) correlations between uncertainty metrics and FActScore. \textbf{Bold} indicates the highest correlation, and \underline{underlined} indicates the second highest in each row. Values marked with `-' indicate inapplicable white-box metrics for API-based models.}
\label{tab:main_results}
\end{table*}

Table~\ref{tab:main_results} presents the performance of our proposed AGSC method compared to various uncertainty
quantification (UQ) baselines. AGSC achieves the highest correlation with FActScore in most experimental settings,
consistently outperforming existing baselines. Among the LUQ variants, LUQ-Atomic generally outperforms the sentence-level
LUQ, confirming that fine-grained verification is crucial. AGSC further advances this by dynamically selecting between
sentence and atomic granularity, avoiding the computational redundancy of full atomic decomposition while maintaining
high precision. In addition, AGSC introduces GMM-based \emph{soft semantic clustering} to obtain theme-aware weights for
aggregation, which mitigates the impact of minor/noisy parts when long responses mix multiple semantic themes.

While all UQ methods show a performance drop on the more challenging LongFact dataset due to its diverse topics and complex reasoning paths, AGSC maintains robust performance. For instance, on Llama3-70B, baseline methods struggle significantly (e.g., SE PCC is -0.067), whereas AGSC maintains a SCC of -0.229, demonstrating its robustness under topic heterogeneity and structural variance in long-context generation.

\subsection{Ablation Study}
\label{sec:ablation}

\begin{table}[t]
\centering
\caption{Ablation study of AGSC using \textbf{GPT-4.1-mini}. We report Pearson (PCC) and Spearman (SCC) correlations between uncertainty and FActScore. \textbf{Bold} indicates the best performance.}
\label{tab:ablation_study}
\resizebox{1\linewidth}{!}{%
\begin{tabular}{llcccc}
\toprule
\multirow{2}{*}{\textbf{Section}} & \multirow{2}{*}{\textbf{Variant}} & \multicolumn{2}{c}{\textbf{BIO}} & \multicolumn{2}{c}{\textbf{LongFact}} \\
\cmidrule(lr){3-4} \cmidrule(lr){5-6}
 & & PCC & SCC & PCC & SCC \\
\midrule
 & \textbf{AGSC} & \textbf{-0.708} & \textbf{-0.665} & \textbf{-0.370} & \textbf{-0.461} \\
\midrule
\multirow{3}{*}{Sec~\ref{sec:adaptive-granularity}} & w/o Adap. & -0.512 & -0.520 & -0.292 & -0.461 \\
 & w/ NG & -0.540 & -0.540 & -0.201 & -0.295 \\
 & w/ NW & -0.564 & -0.514 & -0.198 & -0.257 \\
\midrule
\multirow{2}{*}{Sec~\ref{sec:gmm-alignment}} & w/o Clustering. & -0.679 & -0.622 & -0.204 & -0.307 \\
 & w/ K-Means & -0.531 & -0.462 & -0.229 & -0.361 \\
\bottomrule
\end{tabular}%
}
\end{table}

\paragraph{Impact of Adaptive Granularity.}
We investigated the necessity of our adaptive granularity strategy by comparing it with three variants: (1) \textbf{w/o Adap.}: completely removing the adaptive mechanism; (2) \textbf{w/ NG}: assigning a fixed uncertainty score of $0.5$ to sentences that should have been skipped; and (3) \textbf{w/ NW}: incorporating the neutral probability into the uncertainty calculation with a weight of $0.5$.

As shown in Table~\ref{tab:ablation_study}, removing adaptive granularity (\textit{w/o Adap.}) leads to a significant performance drop on both datasets. This confirms that simply ignoring or uniformly processing neutral sentences is insufficient. Furthermore, both heuristic strategies (\textit{w/ NG} and \textit{w/ NW}) underperform compared to AGSC. This indicates that neutrality in NLI is not a uniform signal of uncertainty or irrelevance. By adaptively distinguishing between \textit{irrelevance tendency} (triggering Skip) and \textit{uncertainty tendency} (triggering Decomposition), AGSC effectively filters noise while retaining verifiable details, which simple numerical heuristics cannot achieve.

\paragraph{Impact of Semantic Clustering.}
We evaluated the effectiveness of the semantic clustering module by removing it (\textbf{w/o Clustering.}) or replacing the
soft GMM clustering with hard K-Means clustering (\textbf{w/ K-Means}).

As shown in Table~\ref{tab:ablation_study}, removing semantic clustering leads to a clear decline in performance,
particularly on LongFact (PCC drops from -0.370 to -0.204). This is expected, as LongFact exhibits stronger topic
heterogeneity and structural variance, where naive uniform aggregation is more likely to be dominated by minor/noisy
portions. Additionally, using K-Means performs significantly worse than the GMM approach (-0.531 vs. -0.708 on BIO),
suggesting that semantic boundaries in long-form generation are often ambiguous. Hard clustering forces each sentence into
a single topic and discards uncertainty in theme membership, whereas GMM's soft assignment preserves probabilistic
memberships, enabling more stable theme-aware aggregation and more robust uncertainty estimation. A qualitative case study
illustrating the learned cluster memberships and their impact on aggregation is provided in Appendix~\ref{app:case_study}.

\subsection{Analysis}
\label{sec:further_analysis}

\subsubsection{Robustness Analysis}

\paragraph{Impact of NLI Backbone.}
\begin{table}[t]
\centering
\caption{Impact of different NLI backbone models on AGSC performance.}
\label{tab:nli_backbone_ablation}
\resizebox{1\linewidth}{!}{%
\begin{tabular}{lcccc}
\toprule
\multirow{2}{*}{\textbf{NLI Backbone}} & \multicolumn{2}{c}{\textbf{BIO}} & \multicolumn{2}{c}{\textbf{LongFact}} \\
\cmidrule(lr){2-3} \cmidrule(lr){4-5}
 & PCC & SCC & PCC & SCC \\
\midrule
DeBERTa-v3-large-mnli (Default) & -0.708 & -0.665 & -0.370 & -0.461 \\
\midrule
GPT-5 & \textbf{-0.735} & \textbf{-0.692} & \textbf{-0.405} & \textbf{-0.498} \\
RoBERTa-large-mnli & -0.685 & -0.642 & -0.358 & -0.449 \\
BART-large-mnli & -0.668 & -0.625 & -0.345 & -0.436 \\
DeBERTa-base-mnli & -0.632 & -0.594 & -0.328 & -0.415 \\
\bottomrule
\end{tabular}%
}
\end{table}

We analyze the sensitivity of AGSC to different NLI models in Table~\ref{tab:nli_backbone_ablation}. The results indicate a positive correlation between the capability of the NLI model and the final UQ performance. Utilizing \textbf{GPT-5} as the NLI backbone yields the best performance, suggesting that the upper bound of our method can be further extended with stronger NLI systems. However, considering the trade-off between cost and performance, our default choice, \textbf{DeBERTa-v3-large}, offers a balanced solution, significantly outperforming smaller models like RoBERTa-large and DeBERTa-base.

\paragraph{Impact of Prompt Granularity.}
\begin{figure}[t]
    \centering
    \includegraphics[width=1\linewidth]{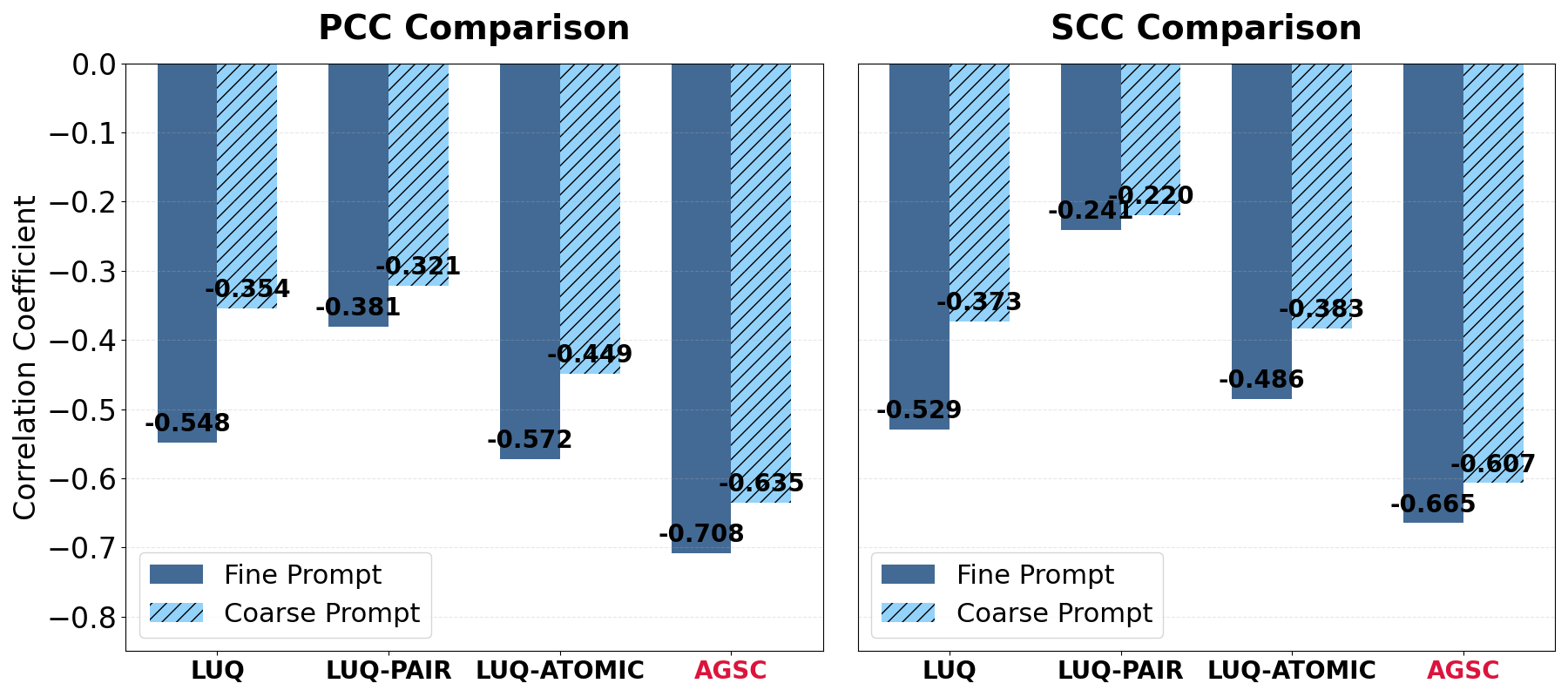}
    \caption{Robustness analysis of AGSC compared to LUQ variants across different prompt granularities.}
    \label{fig:robustness_analysis}
\end{figure}

We investigate how prompt specificity affects uncertainty estimation, as illustrated in Figure~\ref{fig:robustness_analysis}. We compare \textit{Fine-grained Prompts} (e.g., ``Tell me a short bio... Begin with birth... Include education...'') against \textit{Coarse-grained Prompts} (e.g., ``Tell me a short bio...''). 
Experiments show that fine-grained prompts generally lead to higher UQ correlations by reducing aleatoric uncertainty in writing style. However, AGSC demonstrates greater resilience to coarse prompts compared to baselines, benefiting from its GMM-based soft semantic clustering, which provides theme-aware aggregation weights and thus is less sensitive to structural variance inherent in open-ended generation.

\subsubsection{Parameter Sensitivity}

\paragraph{Sensitivity Analysis of Adaptive Threshold.}
\label{sec:sensitivity}

\begin{figure}[t]
    \centering
    \includegraphics[width=1.0\linewidth]{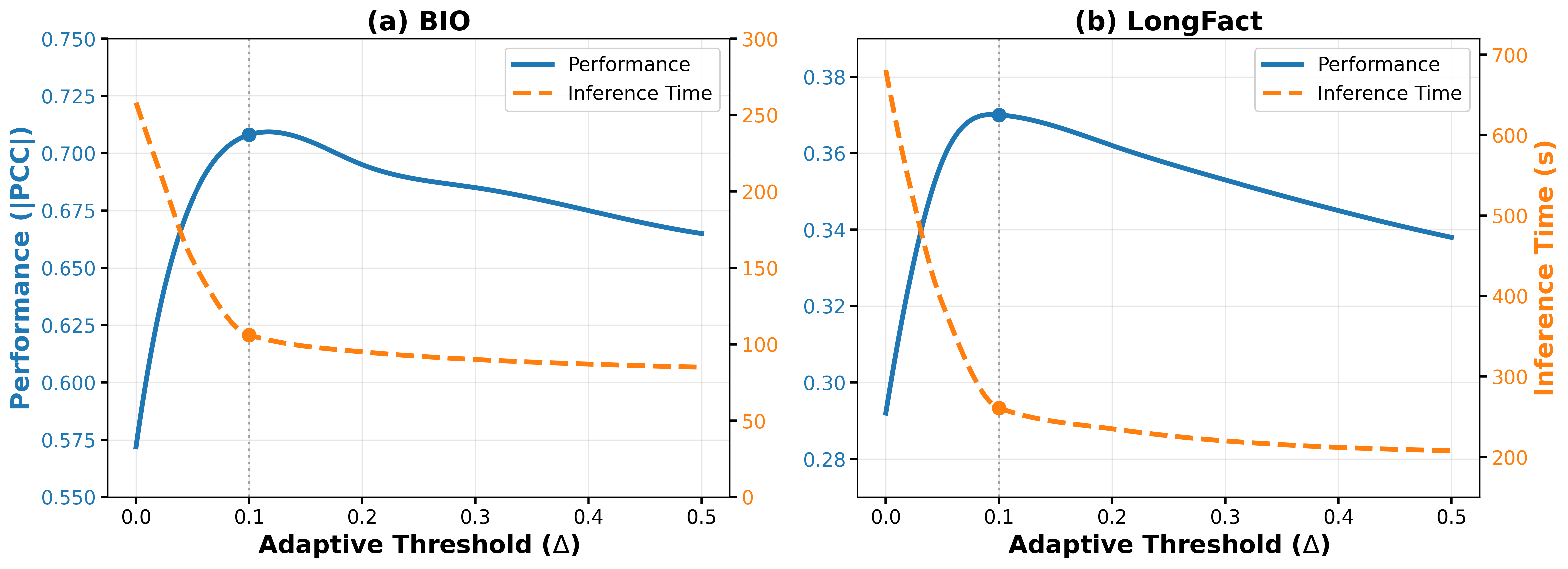}
    \caption{Sensitivity analysis of the adaptive threshold $\Delta$ on BIO and LongFact. 
Left y-axis (solid): Performance ($|$PCC$|$); right y-axis (dashed): Inference Time. 
For both datasets, the optimal trade-off is achieved at $\Delta = 0.1$.}
    \label{fig:threshold_sensitivity}
\end{figure}

As shown in Figure~\ref{fig:threshold_sensitivity}, the optimal threshold is consistently achieved at $\Delta=0.1$ on both BIO and LongFact.
On BIO, decreasing $\Delta$ leads to over-decomposition and unnecessary computation, while increasing $\Delta$ gradually discards useful uncertainty signals.
A similar trend is observed on the more heterogeneous LongFact benchmark, confirming that the adaptive routing threshold generalizes beyond the relatively homogeneous BIO setting.

\paragraph{Impact of Response Quantity.}
\begin{figure}[t]
    \centering
    \includegraphics[width=0.95\linewidth]{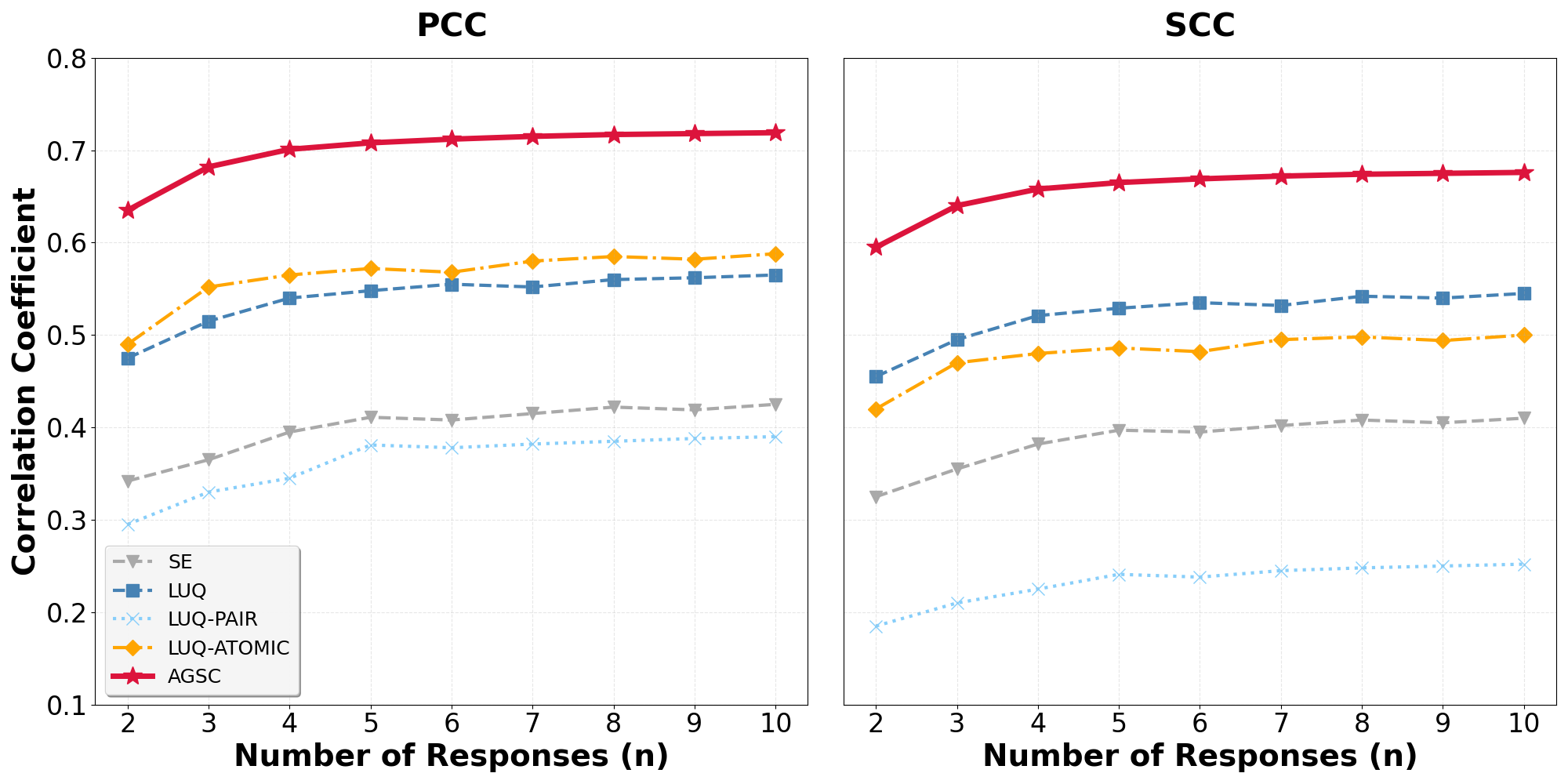}
    \caption{Impact of the number of sampled responses ($n$) on uncertainty estimation performance. The left and right plots show Pearson (PCC) and Spearman (SCC) correlations, respectively.}
    \label{fig:k_ablation}
\end{figure}

Figure~\ref{fig:k_ablation} plots the UQ performance against the number of sampled responses $n$. As expected, performance improves with $n$. Notably, AGSC consistently outperforms other baselines across all sample sizes (from $n=2$ to $10$), demonstrating superior stability.

\subsubsection{Efficiency and Mechanism Analysis}

\paragraph{Time Efficiency Analysis.}
\begin{figure}[t]
    \centering
    \includegraphics[width=0.75\linewidth]{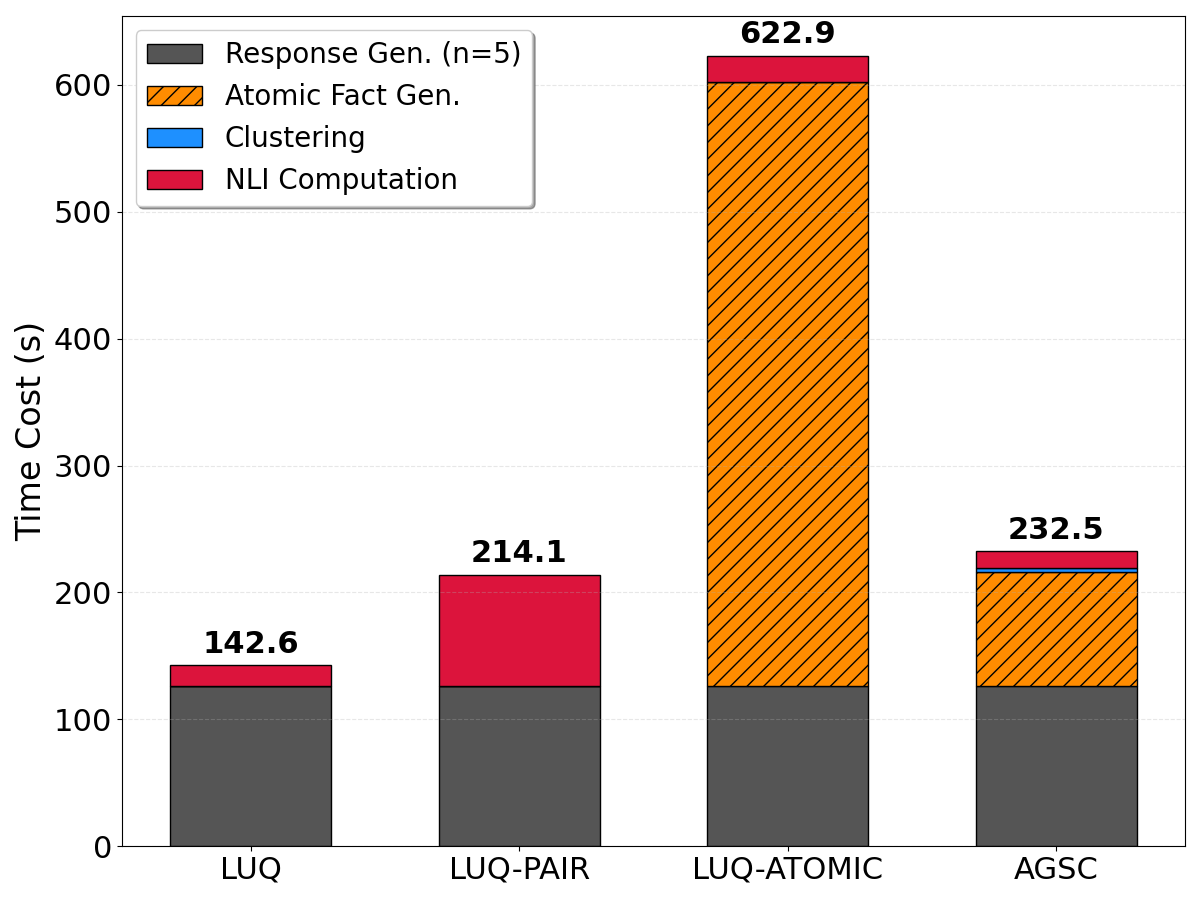}
    \caption{Time consumption breakdown of different uncertainty estimation methods ($n$=5).}
    \label{fig:time_breakdown}
\end{figure}

\begin{figure}[t]
    \centering
    \includegraphics[width=0.75\linewidth]{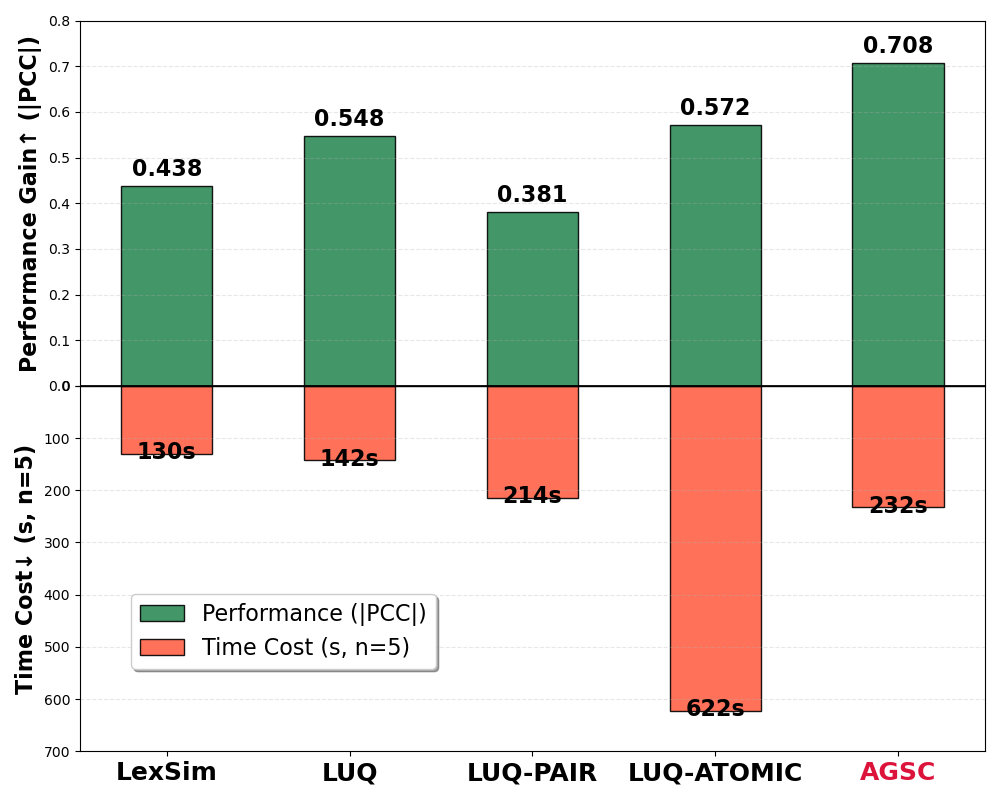}
    \caption{Cost-benefit analysis comparing performance gain versus time cost.}
    \label{fig:cost_benefit_analysis}
\end{figure}

Efficiency is a critical bottleneck for fine-grained semantic UQ.
Figure~\ref{fig:time_breakdown} presents a fine-grained breakdown of computational latency.
For \textbf{LUQ}, the total time is dominated by response generation $T_{\text{gen}}$.
In \textbf{LUQ-Pair}, the quadratic complexity causes $T_{\text{nli}}$ to become a new bottleneck.
\textbf{LUQ-Atomic} introduces an additional atomic fact generation step $T_{\text{atom}}$. Since it indiscriminately decomposes every sentence, $T_{\text{atom}}$ becomes the overwhelming dominant cost.
In contrast, \textbf{AGSC} introduces a strategic optimization. Although it adds a clustering step $T_{\text{cluster}}$, this overhead is negligible. Crucially, by filtering out irrelevant neutral sentences, AGSC drastically reduces the volume of text requiring decomposition, thereby significantly cutting down $T_{\text{atom}}$.

Figure~\ref{fig:cost_benefit_analysis} illustrates the Time-Performance Trade-off, where AGSC occupies the optimal position on the Pareto frontier. Crucially, AGSC achieves superior performance compared to LUQ-Atomic while reducing the total inference time by approximately 60\%, primarily by bypassing the decomposition of irrelevant neutral sentences.

\paragraph{NLI Distribution and Adaptive Effectiveness.}
\begin{figure}[t]
    \centering
    \includegraphics[width=0.85\linewidth]{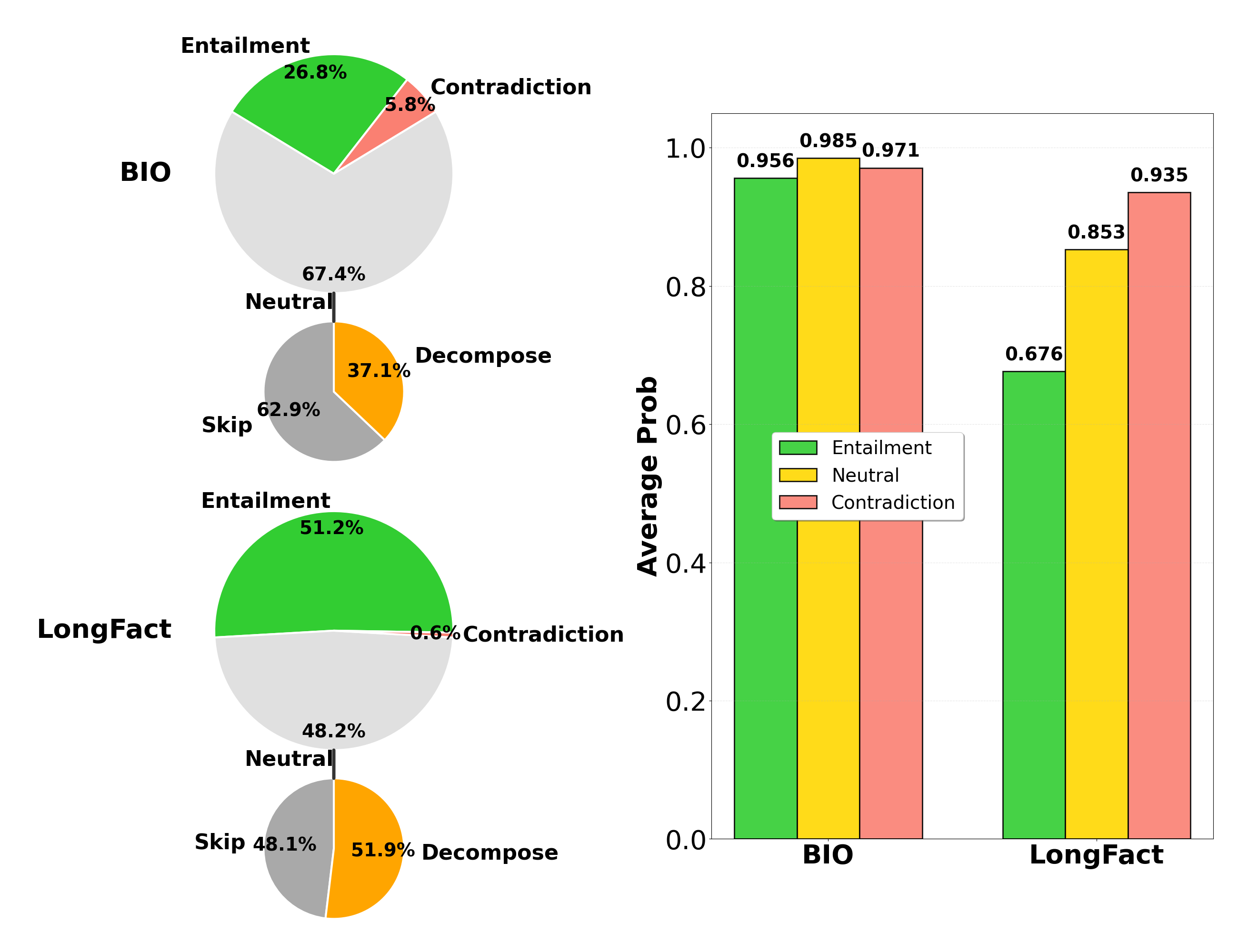}
    \caption{Analysis of NLI Distribution.}
    \label{fig:nli_granularity_analysis}
\end{figure}

To understand the necessity of our adaptive mechanism, we analyze the NLI statistics in Figure~\ref{fig:nli_granularity_analysis}. \textbf{Figure~\ref{fig:nli_granularity_analysis} (Left)} shows that the \textit{Neutral} category constitutes a massive portion (40-70\%) of the generated sentences. Within these neutral samples, $\sim$30-50\% exhibit an \textit{Uncertainty Tendency} that necessitates decomposition. Ignoring such a large segment leads to significant information loss.

\textbf{Figure~\ref{fig:nli_granularity_analysis} (Right)} presents the average probability of the corresponding category. We observe a distinct \textbf{probability saturation} on the BIO dataset: the NLI model assigns probabilities $>0.95$ even for Neutral and Contradiction labels. This implies that the absolute value of $P_N$ is insufficient to distinguish between Irrelevance and Uncertainty tendencies.
AGSC addresses this by utilizing the \textit{Entailment-Contradiction Gap} ($\Delta$). By distinguishing tendencies based on polarity difference, AGSC triggers atomic decomposition for only $\sim$25\% of the potential candidates compared to LUQ-Atomic. This selective granularity allows AGSC to achieve state-of-the-art performance with significantly reduced computational overhead.

\section{Conclusion}
\label{sec:conclusion}
We propose \textbf{AGSC} to address efficiency and topic heterogeneity challenges in long-form UQ. By integrating an \textbf{Adaptive Granularity Strategy} for selective decomposition with \textbf{GMM-based Semantic Clustering} for theme-aware aggregation, AGSC ensures stable uncertainty estimates. Experiments on BIO and LongFact demonstrate that AGSC achieves SOTA performance while reducing inference latency by $\sim$60\% compared to full atomic decomposition.

\section*{Limitations}
\label{sec:limitations}

\paragraph{Dependency on NLI Calibration.}
The core of our \textit{Adaptive Granularity} mechanism relies on the NLI model's ability to accurately classify the ``Neutral'' category. If the NLI model is miscalibrated---specifically, if it misclassifies subtle contradictions as neutral (false negatives) or fails to detect the entailment relationship in paraphrased texts---the adaptive trigger will fail. For instance, in highly technical domains (e.g., medical or legal texts), general-purpose NLI models (like DeBERTa-v3-trained on MNLI) often lack the domain knowledge to distinguish between a \textit{factual contradiction} and a \textit{neutral supplement}, potentially causing AGSC to incorrectly skip hallucinated content.

\paragraph{Vulnerability to ``Echo Chamber'' Hallucinations.}
Like all consistency-based methods, AGSC assumes that truth converges while hallucinations diverge. However, for certain widespread misconceptions or when the LLM has been poisoned with specific incorrect data during pre-training, the model may consistently generate the same hallucination across all $K$ samples. In such cases of \textit{systematic error}, high semantic consistency will result in a low uncertainty score, misleadingly indicating high factuality. AGSC measures \textit{consensus}, not absolute \textit{truth}.

\paragraph{Structural Constraints.} AGSC is optimized for fact-heavy, descriptive texts (e.g., biographies, clinical notes). It may struggle with content where the ``unit of truth'' is not sentence-based, such as \textbf{mathematical reasoning} or \textbf{code generation}. In these cases, GMM clustering on sentence embeddings may break the logical flow, and NLI models are ill-equipped to verify logic-based entailment.

\paragraph{Linguistic Constraints.}
Our experiments are conducted exclusively in English using English-centric embedding (GTE) and NLI models. The effectiveness
of GMM-based \emph{semantic clustering} relies on the quality of the embedding space. In low-resource languages where
embedding models may exhibit anisotropy or poor semantic separability, the learned clusters (and thus the theme-aware
aggregation weights) can become unstable, which may degrade the robustness of the final uncertainty estimate.

\section*{Ethics Statement}
\label{sec:ethics}

\paragraph{Reliability vs. Truth.}
Users must be aware that a low uncertainty score computed by AGSC indicates that the model is \textit{confident and consistent}, but not necessarily \textit{correct}. Deploying this metric in high-stakes decision-making systems (e.g., automated medical diagnosis or legal advice) carries the risk of \textit{automation bias}, where users might over-trust the system based on a high consistency score derived from shared hallucinations.

\paragraph{Bias Propagation from Backbone Models.}
AGSC leverages NLI models and embedding models that are known to inherit social biases (e.g., gender, race, or religious stereotypes) from their training data. If an NLI model exhibits bias (e.g., consistently classifying premises involving certain demographics as ``contradictions'' or ``neutral'' without logical basis), AGSC will propagate this bias into the final uncertainty score. This could lead to systematically higher uncertainty estimates for content related to marginalized groups.

\paragraph{Artifacts and Licensing.}
In this study, we utilize several open-source scientific artifacts, including the Llama-3 , Qwen-2.5 , and DeBERTa-v3 models, as well as the BIO and LongFact  datasets. All artifacts are used in strict accordance with their respective original licenses. Our proposed AGSC framework and its implementation code are provided in the supplementary materials.

\paragraph{AI Usage Statement.} AI assistants were primarily employed for language polishing and as an auxiliary tool for generating specific code modules within the AGSC framework. It is important to emphasize that all AI-generated content and code were rigorously reviewed and manually verified by the authors to ensure technical accuracy, integrity, and the absence of bias. The final manuscript and implementation represent the original intellectual contributions of the authors.


\bibliography{anthology,custom}
\bibliographystyle{acl_natbib}

\appendix

\section{Adaptive Soft Clustering Algorithm}
\label{app:clustering_algorithm}

\begin{algorithm}[h]
\caption{Adaptive Soft Clustering with UMAP and BIC}
\label{alg:adaptive_clustering}
\begin{algorithmic}[1]
\REQUIRE 
    Generated responses $\mathcal{R} = \{r_1, \dots, r_n\}$; 
    Embedding model $\mathcal{M}_{\text{emb}}$; 
    UMAP target dimension $D' = 32$;
    BIC improvement threshold $\epsilon = 0.01$;
    Global max clusters $K_{\text{limit}} = 15$.
\ENSURE 
    Soft membership matrix $P \in \mathbb{R}^{N \times K}$, Cluster centers $\mu$.

\STATE \textbf{Step 1: Preprocessing \& Embedding}
\STATE Split all responses into a global sentence set $\mathcal{S} = \{s_1, \dots, s_N\}$, where $N = \sum |r_i|$.
\STATE Compute embeddings: $X \leftarrow \mathcal{M}_{\text{emb}}(\mathcal{S})$, where $X \in \mathbb{R}^{N \times D}$.

\STATE \textbf{Step 2: Dimensionality Reduction}
\IF{$D > D'$}
    \STATE Apply UMAP: $X_{\text{reduced}} \leftarrow \text{UMAP}(X, \text{n\_components}=D')$.
\ELSE
    \STATE $X_{\text{reduced}} \leftarrow X$.
\ENDIF

\STATE \textbf{Step 3: Heuristic $K_{\max}$ Selection}
\STATE Calculate density constraint: $K_{\text{density}} \leftarrow \max(2, \lfloor N / 3 \rfloor)$.
\STATE Calculate logarithmic constraint: $K_{\text{log}} \leftarrow \lfloor \log_2(N) \rfloor + 1$.
\STATE Determine search limit:
\begin{equation*}
    K_{\max} \leftarrow \min(K_{\text{limit}}, K_{\text{density}}, K_{\text{log}})
\end{equation*}

\STATE \textbf{Step 4: BIC-based GMM Selection}
\STATE Initialize $BIC_{\text{last}} \leftarrow \infty, \text{BestModel} \leftarrow \text{None}$.
\FOR{$k = 2$ \TO $K_{\max}$}
    \STATE Fit GMM with $k$ components on $X_{\text{reduced}}$: $\mathcal{G}_k$.
    \STATE Calculate $BIC_{\text{current}} \leftarrow \text{BIC}(\mathcal{G}_k, X_{\text{reduced}})$.
    \STATE \textit{// Check relative improvement}
    \IF{$\text{BestModel}$ is None \OR $BIC_{\text{current}} < BIC_{\text{last}} - \epsilon \cdot |BIC_{\text{last}}|$}
        \STATE $\text{BestModel} \leftarrow \mathcal{G}_k$.
        \STATE $BIC_{\text{last}} \leftarrow BIC_{\text{current}}$.
    \ELSE
        \STATE \textbf{break} \textit{// Stop if improvement is negligible}
    \ENDIF
\ENDFOR

\STATE \textbf{Step 5: Final Projection}
\STATE Compute membership probabilities $P$ using $\text{BestModel}$ on $X_{\text{reduced}}$.
\STATE \textbf{return} $P, \text{BestModel}.\mu$.
\end{algorithmic}
\end{algorithm}
To address the varying number of sentences and semantic densities in long-form generation, we designed a GMM-based adaptive clustering algorithm. It integrates UMAP dimensionality reduction, a heuristic estimation for the maximum number of clusters ($K_{\max}$), and BIC-based model selection. The detailed procedure is outlined in Algorithm~\ref{alg:adaptive_clustering}.

\section{Qualitative Case Study}
\label{app:case_study}

\begin{figure}[h]
\centering
\begin{tcolorbox}[colback=gray!5!white, colframe=gray!75!black, title=Adaptive Granularity Decisions]
\textbf{Case 1: The ``Skip'' Decision (Irrelevance Tendency)} \\
\textit{\textbf{Sentence:}} ``In summary, 'Black Swan' is a layered, dark exploration of the pursuit of perfection...'' \\
\textit{\textbf{Context:}} This sentence appears at the end of a response as a concluding remark. \\
\textit{\textbf{NLI Analysis:}} When compared to other factual responses, this sentence yields a \textbf{Neutral} prediction because it expresses a subjective summary not explicitly present in other samples. \\
\textit{\textbf{Adaptive Trigger:}} The model detects a low Entailment-Contradiction gap ($\Delta \le 0.1$), correctly identifying this as phatic/opinionated content. \\
\textit{\textbf{Action:}} \textbf{\textsc{Skip}}. This prevents the model from hallucinating atomic facts for subjective opinions, reducing noise.

\vspace{0.8em}
\hrule
\vspace{0.8em}

\textbf{Case 2: The ``Decompose'' Decision (Uncertainty Tendency)} \\
\textit{\textbf{Sentence:}} ``Directed by Darren Aronofsky and co-written by him and Mark Heyman.'' \\
\textit{\textbf{Comparison Target:}} Another response lists writers as ``Mark Heyman, Andres Heinz, John McLaughlin'', without mentioning Aronofsky as a co-writer. \\
\textit{\textbf{NLI Analysis:}} The sentence contains mixed veracity. ``Directed by Aronofsky'' is Entailment, but ``co-written by him'' is not supported by the target response (\textbf{Neutral}). \\
\textit{\textbf{Adaptive Trigger:}} The conflict creates a high gap ($\Delta > 0.1$). \\
\textit{\textbf{Action:}} \textbf{Decompose} into atomic facts:
\begin{enumerate}
    \item \textit{\textbf{Directed by Darren Aronofsky}} $\rightarrow$ \textcolor{green}{Supported}
    \item \textit{\textbf{Co-written by Darren Aronofsky}} $\rightarrow$ \textcolor{red}{Contradiction/Neutral}
    \item \textit{\textbf{Co-written by Mark Heyman}} $\rightarrow$ \textcolor{green}{Supported}
\end{enumerate}
\end{tcolorbox}
\caption{Examples of the Skip vs. Decompose logic in AGSC's adaptive granularity module.}
\label{fig:adaptive_granularity_case}
\end{figure}

\begin{figure}[h]
\centering
\begin{tcolorbox}[colback=gray!5!white, colframe=gray!75!black, title=Semantic Clustering for Theme-aware Weighting] \textbf{Target Sentence ($s$):} `The film stars Natalie Portman as Nina Sayers...'' (from Response 1) \\ \textbf{Without Clusteringt (Standard LUQ):} \\ $s$ is compared against \textit{all} sentences in Response 2: \begin{itemize} \item vs. `Black Swan is a horror film...'' $\rightarrow$ Neutral (Noise) \item vs. `The film stars Natalie Portman...'' $\rightarrow$ \textbf{Entailment} \item vs. `Plot Summary: The film follows...'' $\rightarrow$ Neutral (Noise) \end{itemize} \textit{\textbf{Result:}} The strong entailment signal is diluted by multiple Neutral scores from irrelevant sections (Plot, Intro). \vspace{0.5em} \\
\textbf{With GMM Clustering (AGSC):} \\ $s$ is mapped to the `Cast \& Characters'' cluster. \begin{itemize} \item \textbf{High Weight:} `The film stars Natalie Portman...'' (Cluster Match) \item \textbf{Low Weight:} `Black Swan is a horror film...'' (Different Cluster) \end{itemize} \textit{\textbf{Result:}} The uncertainty score is derived primarily from the semantically relevant counterpart, resulting in a high-confidence \textbf{Entailment} score. \end{tcolorbox}
\caption{Comparison of NLI targets before and after GMM semantic clustering.}
\label{fig:alignment_case}
\end{figure}

\begin{figure}[h]
\centering
\begin{tcolorbox}[colback=gray!5!white, colframe=gray!75!black, title=Clustering Failure Analysis]
\textit{\textbf{Example:}} ``The film explores themes of ambition and duality.'' \\
\textit{\textbf{Scenario:}} If sentence embeddings do not separate ``themes'' from ``plot summary'' cleanly, the soft
memberships may spread across multiple clusters or assign low mass to the intended theme. \\
\textit{\textbf{Consequence:}} Theme-aware aggregation may underweight this sentence (treating it as minor context),
leading to a higher uncertainty estimate even when the statement is factual. \\
\textit{\textbf{Result:}} This is a clustering/representation limitation: the method can mis-estimate uncertainty due to
unstable theme weights, even though the NLI comparison target (whole reference responses) remains unchanged.
\end{tcolorbox}
\caption{Analysis of a GMM clustering failure due to cross-cluster thematic ambiguity.}
\label{fig:failure_mode_case}
\end{figure}

In this section, we provide a qualitative analysis of real-world examples from the \textbf{LongFact} dataset (Topic: ``Black Swan'' movie) to demonstrate the effectiveness of AGSC's components and discuss their limitations.

\subsection{Adaptive Granularity in Action}

Standard sentence-level methods treat all ``Neutral'' NLI predictions equally. However, our Adaptive Granularity strategy distinguishes between ``Irrelevant'' content (to be skipped) and ``Uncertain/Complex'' content (to be decomposed) based on the entailment-contradiction gap ($\Delta$). Figure~\ref{fig:adaptive_granularity_case} illustrates how these decisions are made.

\subsection{Impact of GMM Semantic Clustering}
Long-form responses often mix multiple semantic themes. We illustrate how GMM-based soft clustering provides
theme-aware weights that stabilize aggregation under topic heterogeneity (Figure~\ref{fig:alignment_case}).

\subsection{Failure Mode Analysis: Clustering Instability}
While semantic clustering improves robustness, it is not infallible. Figure~\ref{fig:failure_mode_case} describes a
typical failure mode caused by ambiguous theme boundaries in the embedding space.
\end{document}